\documentclass[letterpaper]{article}
\usepackage{iccc}

\usepackage{url}  
\usepackage{cleveref}
\usepackage[font=small,labelfont=bf]{caption}
\usepackage{graphicx}  
\usepackage{subfigure}
\usepackage{etoolbox,lipsum}
\usepackage{times}
\usepackage{helvet}
\usepackage{courier}

\DeclareRobustCommand{\VAN}[3]{#2} 

\title{Shallow Art: Art Extension Through Simple Machine Learning}

\author{Kyle Robinson, Dan Brown\\
Cheriton School of Computer Science\\
Unversity of Waterloo\\
Waterloo, ON N2L 3G1  Canada\\
\{kyle.robinson, dan.brown\}@uwaterloo.ca\\
}
\begin{document} 
\maketitle
\begin{abstract}
\begin{quote}
Shallow Art presents, implements, and tests the use of simple single-output classification and regression models for the purpose of art generation. Various machine learning algorithms are trained on collections of computer generated images, artworks from Vincent van Gogh, and artworks from Rembrandt van Rijn. These models are then provided half of an image and asked to complete the missing side. The resulting images are displayed, and we explore implications for computational creativity.
\end{quote}
\end{abstract}

\section{Introduction}
The use of Machine Learning algorithms as assistive tools for the creation of visual and auditory artworks is an ever growing area of research~(\citeauthor{magenta}). In addition, machine learning methodologies--especially those based on neural networks--have been used in the creation of wholly new artworks~\cite{deep_art}. The creation and use of such tools encourages discussions at the boundary between computational creativity and creativity support.

Many machine learning computational creativity applications have thus far used neural networks. While these methods have lead to extremely interesting and thought-provoking artwork~(\citeauthor{quasi,obvious}), their analysis from a creative perspective is complicated by an inability to interpret \textit{why} the model acted as it did. This black-box effect can in some ways parallel our understanding of human creativity but differs in many ways, especially as neural networks are meticulously designed to fulfill their specific purpose.

In contrast to neural network based approaches of algorithmic art generation, Shallow Art applies more classical Machine Learning algorithms and methodologies to the problem of art generation. More specifically, Shallow Art applies computationally efficient classification and regression algorithms to the problem of art extension. We have developed a methodology that allows any single-output classification or regression algorithm to be trained on a dataset of images and then \textit{complete} a partially provided image. We conclude by presenting outputs from various models trained using a number of different types of training images.

\section{Related Work}
Work on machine learning based creativity support is rich in breadth and depth. One significant source of recent research is the Magenta research division of Google Brain~(\citeauthor{magenta}). In the area of music generation, their work includes a model which creates piano performances featuring expressive changes in tempo and dynamics, as well as a musical counterpoint generator using a specifically-designed convolutional neural network~\cite{piano_create,piano_counterpoint}. Additionally, Magenta tools transcribe polyphonic music, and synthesize sounds for music production~\cite{piano_trans,synth}. In visual art, Magenta has developed a recurrent neural network able to create very simple stick drawings~\cite{sketches}. The network is trained on human drawings with a provided single classification and can generate drawings with or without a provided topic. These drawings are extremely simple in nature; consisting of just a few simple lines and shapes. 

One interesting computational art generation research paper similar to Shallow Art attempts to learn the artistic style of one picture and transfer it to another as if the second image was created in the style of the first~\cite{deep_art}. Where all of these research papers differ from Shallow Art is in their use of black-box or grey-box neural networks for training and generation. Conversely, Shallow Art prioritizes and enables the use of easily understandable and traceable machine learning algorithms such as decision trees and seperating hyperplanes, with each pixel of the resultant project treated independently.

\section{Methodology}

\subsection{Image Source}

The data for this project is sourced through a combination of black-and-white computer-generated images of varying complexities, colored computer generated images, and a collection of artworks created by Vincent van Gogh and Rembrandt van Rijn. In the black-and-white data set each pixel of each image is either white or black. In the coloured image data sets each pixel is represented by a 3-tuple \((Red ,Green, Blue)\) of integers ranging in value from 0-255. 

\setlength{\fboxsep}{0pt}
\newlength{\imagewidth}
\begin{figure*}%
\centering
\subfigure[][]{%
\label{fig:horizontal_input}%
\settowidth{\imagewidth}{\includegraphics[]{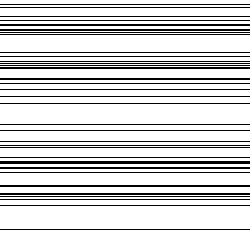}}
\fbox{\fbox{\includegraphics[trim=0 0 0.5\imagewidth{} 0, clip, width = 0.053\linewidth]{assets/horizontal_1.png}}
\fbox{\includegraphics[trim=0.5\imagewidth{} 0 0 0, clip, width = 0.053\linewidth]{assets/horizontal_1.png}}}%
\hspace{0pt}
\settowidth{\imagewidth}{\includegraphics[]{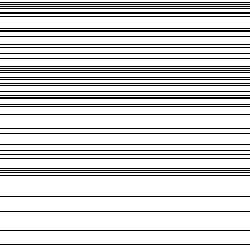}}
\fbox{\fbox{\includegraphics[trim=0 0 0.5\imagewidth{} 0, clip, width = 0.053\linewidth]{assets/horizontal_2.png}}
\fbox{\includegraphics[trim=0.5\imagewidth{} 0 0 0, clip, width = 0.053\linewidth]{assets/horizontal_2.png}}}}%
\hspace{0pt}
\subfigure[][]{%
\label{fig:vertical_input}%
\settowidth{\imagewidth}{\includegraphics[]{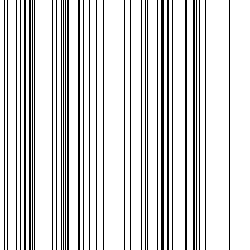}}
\fbox{\fbox{\includegraphics[trim=0 0 0.5\imagewidth{} 0, clip, width = 0.053\linewidth]{assets/vertical_1.png}}
\fbox{\includegraphics[trim=0.5\imagewidth{} 0 0 0, clip, width = 0.053\linewidth]{assets/vertical_1.png}}}%
\hspace{0pt}
\settowidth{\imagewidth}{\includegraphics[]{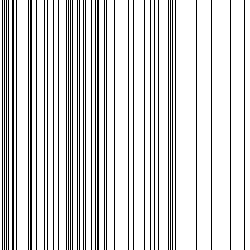}}
\fbox{\fbox{\includegraphics[trim=0 0 0.5\imagewidth{} 0, clip, width = 0.053\linewidth]{assets/vertical_2.png}}
\fbox{\includegraphics[trim=0.5\imagewidth{} 0 0 0, clip, width = 0.053\linewidth]{assets/vertical_2.png}}}}%
\hspace{0pt}
\subfigure[][]{%
\label{fig:circle_input}%
\settowidth{\imagewidth}{\includegraphics[]{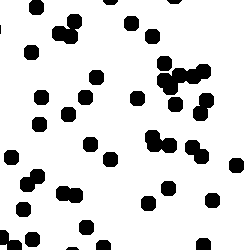}}
\fbox{\fbox{\includegraphics[trim=0 0 0.5\imagewidth{} 0, clip, width = 0.053\linewidth]{assets/circles_1.png}}
\fbox{\includegraphics[trim=0.5\imagewidth{} 0 0 0, clip, width = 0.053\linewidth]{assets/circles_1.png}}}%
\hspace{0pt}
\settowidth{\imagewidth}{\includegraphics[]{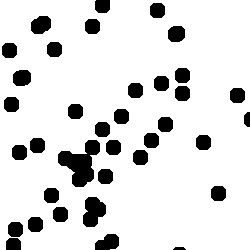}}
\fbox{\fbox{\includegraphics[trim=0 0 0.5\imagewidth{} 0, clip, width = 0.053\linewidth]{assets/circles_2.png}}
\fbox{\includegraphics[trim=0.5\imagewidth{} 0 0 0, clip, width = 0.053\linewidth]{assets/circles_2.png}}}}%
\hspace{0pt}
\subfigure[][]{%
\label{fig:triangle_input}%
\settowidth{\imagewidth}{\includegraphics[]{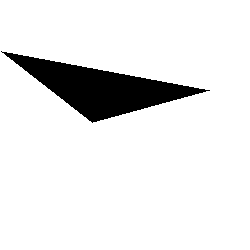}}
\fbox{\fbox{\includegraphics[trim=0 0 0.5\imagewidth{} 0, clip, width = 0.053\linewidth]{assets/poly_1.png}}
\fbox{\includegraphics[trim=0.5\imagewidth{} 0 0 0, clip, width = 0.053\linewidth]{assets/poly_1.png}}}%
\hspace{0pt}
\settowidth{\imagewidth}{\includegraphics[]{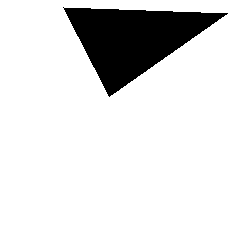}}
\fbox{\fbox{\includegraphics[trim=0 0 0.5\imagewidth{} 0, clip, width = 0.053\linewidth]{assets/poly_2.png}}
\fbox{\includegraphics[trim=0.5\imagewidth{} 0 0 0, clip, width = 0.053\linewidth]{assets/poly_2.png}}}}%
\hspace{0pt}
\subfigure[][]{%
\label{fig:triangle_color_input}%
\settowidth{\imagewidth}{\includegraphics[]{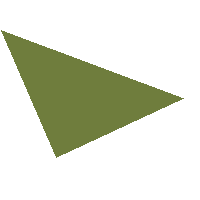}}
\fbox{\fbox{\includegraphics[trim=0 0 0.5\imagewidth{} 0, clip, width = 0.0735\linewidth]{assets/poly_art_1.png}}
\fbox{\includegraphics[trim=0.5\imagewidth{} 0 0 0, clip, width = 0.0735\linewidth]{assets/poly_art_1.png}}}%
\hspace{0pt}
\settowidth{\imagewidth}{\includegraphics[]{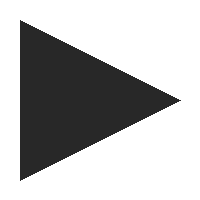}}
\fbox{\fbox{\includegraphics[trim=0 0 0.5\imagewidth{} 0, clip, width = 0.0735\linewidth]{assets/poly_art_2.png}}
\fbox{\includegraphics[trim=0.5\imagewidth{} 0 0 0, clip, width = 0.0735\linewidth]{assets/poly_art_2.png}}}}%
\hspace{0pt}
\subfigure[][]{%
\label{fig:gogh_input}%
\settowidth{\imagewidth}{\includegraphics[]{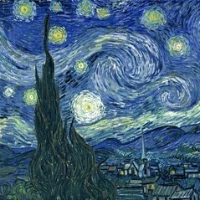}}
\fbox{\fbox{\includegraphics[trim=0 0 0.5\imagewidth{} 0, clip, width = 0.0735\linewidth]{assets/gogh_2.jpg}}
\fbox{\includegraphics[trim=0.5\imagewidth{} 0 0 0, clip, width = 0.0735\linewidth]{assets/gogh_2.jpg}}}%
\hspace{0pt}
\settowidth{\imagewidth}{\includegraphics[]{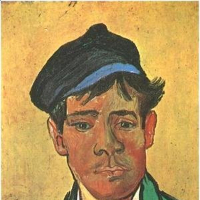}}
\fbox{\fbox{\includegraphics[trim=0 0 0.5\imagewidth{} 0, clip, width = 0.0735\linewidth]{assets/gogh_1.jpg}}
\fbox{\includegraphics[trim=0.5\imagewidth{} 0 0 0, clip, width = 0.0735\linewidth]{assets/gogh_1.jpg}}}}%
\hspace{0pt}
\subfigure[][]{%
\label{fig:rem_input}%
\settowidth{\imagewidth}{\includegraphics[]{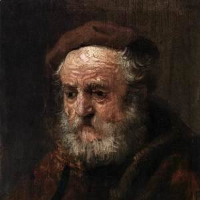}}
\fbox{\fbox{\includegraphics[trim=0 0 0.5\imagewidth{} 0, clip, width = 0.0735\linewidth]{assets/rem_1.jpg}}
\fbox{\includegraphics[trim=0.5\imagewidth{} 0 0 0, clip, width = 0.0735\linewidth]{assets/rem_1.jpg}}}%
\hspace{0pt}
\settowidth{\imagewidth}{\includegraphics[]{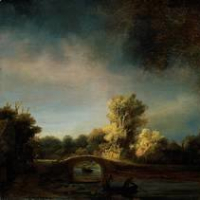}}
\fbox{\fbox{\includegraphics[trim=0 0 0.5\imagewidth{} 0, clip, width = 0.0735\linewidth]{assets/rem_2.jpg}}
\fbox{\includegraphics[trim=0.5\imagewidth{} 0 0 0, clip, width = 0.0735\linewidth]{assets/rem_2.jpg}}}}%
\hspace{0pt}
\caption{\textbf{Image Data.} Two examples each of a selection of image types used to train and test classification and regression models. (a)-(e) were randomly computer generated, whereas (f) and (g) are from collections of van Gogh and Rembrandt works respectively.}%
\label{fig:input}%
\end{figure*}

The computer generated image data sets were created using Python 3 and the Python Imaging Library through the Pillow distribution~(\citeauthor{pillow}). Four arbitrary patterns were selected for initial black and white image testing (\cref{fig:horizontal_input,fig:vertical_input,fig:circle_input,fig:triangle_input}): horizontal lines, vertical lines, triangles, and circles. Each image type eliminates all structure except one abstract concept which is exceedingly easy for a human to perceive. These generated images can also be sorted into two groups: those with unique solutions to the right half (that is, where knowing the left half implies the right half), and those without unique solutions to the right half. Note that as the resolution of the generated images decreases, the approximations of lines, and curves drastically decrease due to rasterization. Black and white images were generated at a resolution of 250x250 pixels and stored in the PNG format. Color images (including van Gogh and Rembrandt works) were generated or cropped to a resolution of 200x200 pixels due to limitations in the van Gogh and Rembrandt data repositories.

\subsubsection{Lines}
 Two types of line-based images were generated: horizontal, and vertical. Each image begins as a square white image, then horizontal or vertical lines one pixel thick are drawn at 50 random locations. Given any horizontal line image's left side the right side has only one correct solution, whereas the vertical line images have no left-to-right continuity. Examples of these images can be seen in~\Cref{fig:horizontal_input,fig:vertical_input}.

\subsubsection{Circles}
Generated circle images consist of a white background with 50 randomly placed black circles. All 50 circles have a diameter of 15 pixels and may be placed at locations which clip with the edge of the image by up to their diameter. Circles are drawn as rasterized approximations of circles. Given any circle image's left side the right side contains some unique solutions (where circles are split in half) and many unknown solutions (the rest). Examples of circle images can be seen in~\Cref{fig:circle_input}

\subsubsection{Triangles}
We generated triangle images with two randomly chosen vertices on the left side and one on the right. These vertices are then connected and filled in order to create a black triangle on a white background. Given any left-side image, predicting the right side is a relatively trivial task of extending the triangles' edges until they converge at the third vertex. Examples of these images can be seen in~\Cref{fig:triangle_input}. We also generated a set of triangle images that were each assigned a random fill colour and placed on a white background. Examples can be seen in~\Cref{fig:triangle_color_input}

\subsubsection{Van Gogh \& Rembrandt}
We obtained Vincent van Gogh and Rembrandt van Rijn works (public domain) by scraping thumbnails from online galleries~(\citeauthor{gogh,rembrandt}). In total we collected 2,298 van Gogh images and 1,097 Rembrandt images. These images consist of finished and unfinished works of differing aspect ratios and sizes. We standardized these data sets by removing duplicates and non-rectangular images and scaling all remaining images to 200x200 pixels as seen in~\Cref{fig:gogh_input,fig:rem_input}. The van Gogh and Rembrandt images pixel data is stored and accessed in the 3-tuple RGB format described earlier.

\begin{figure}[]
    \centering
    \includegraphics[width=0.4\linewidth]{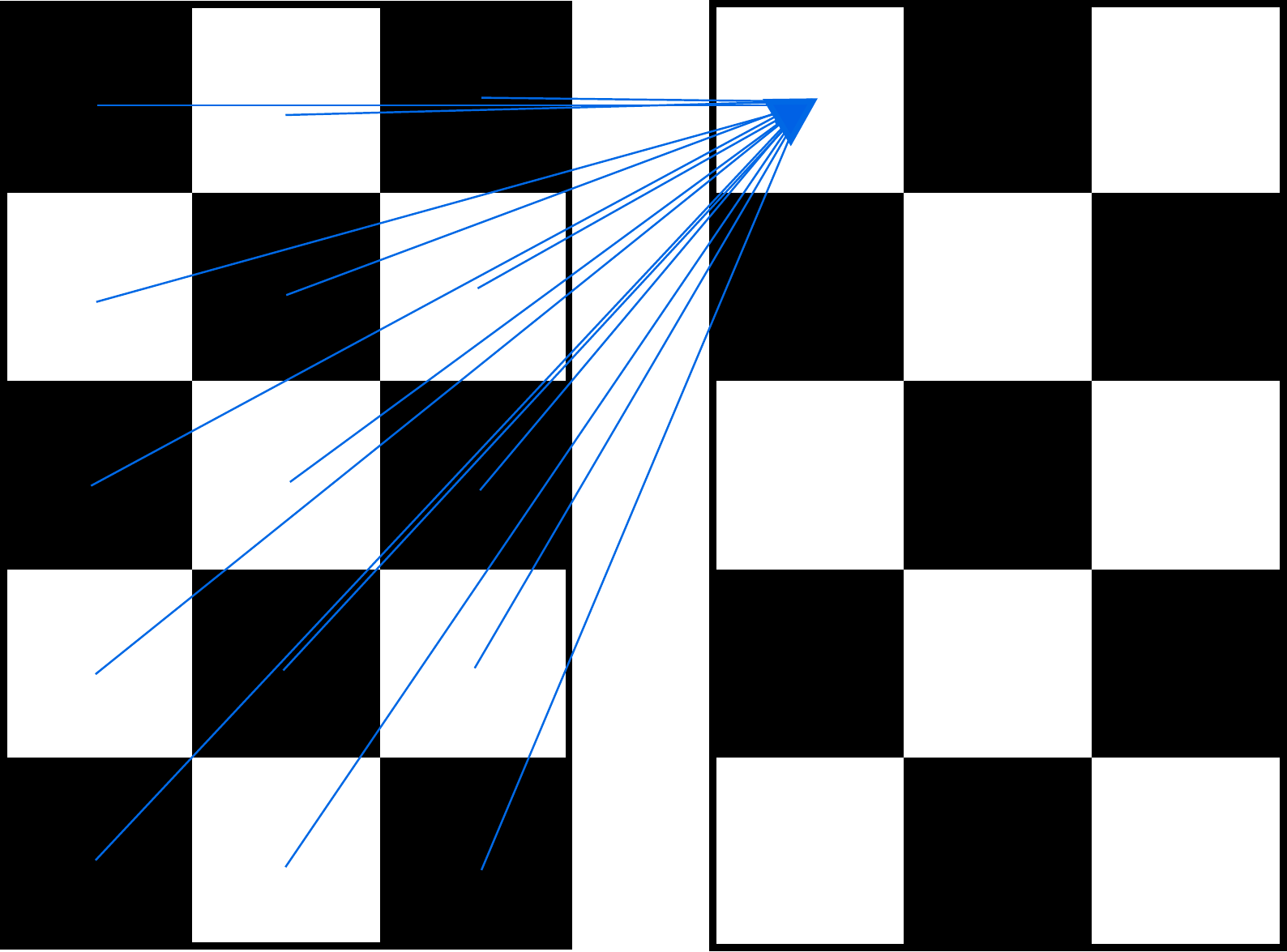}
    \caption{Each pixel on the right is predicted using the values of all pixels on the left (attributes). This process is repeated so that each pixel on the right half of the image is the output of one independent model.}
    \label{fig:checkerboard}
\end{figure}

\subsection{Machine Learning}

The task of binary prediction on the black and white data sets is one of binary classification, whereas the prediction of coloured images is one of regression, since the output RGB values must be between 0 and 255. In order to prepare images for machine learning classification and regression, each image is converted into a flattened one-dimensional array of values. For black and white images, each value is a binary representation of the pixel. For color images, each pixel corresponds to three values in the array, one for each colour shading. Image pixel arrays are split in half so that one half contains all pixels of the left side of the image and the other contains all pixels from the right side. The data array corresponding the left of the image is used as a series of attributes, and the array corresponding to the right side of the image is used as series of labels. In this way the black and white images contain \(\frac{250\times250}{2}=31,250\) training attributes each and the color images contain \(\frac{(200\times200)\times3}{2}=60,000\) attributes each.

As traditional regression and classification tasks involve many attributes and one prediction per data point a method of converting single-output classification and regression models to multi-output is required. One simple and computationally efficient method of doing so is to train one model for each output required as described in~\Cref{fig:checkerboard}. Our implementation is altogether independent: it consumes any classification or regression model and trains one independent model for each of any number of required outputs. Due to the halving of images in this work the total number of trained models required for each image type is equal to the number of attributes. There are thus 31,250 models for black and white images and 60,000 models for color images. We refer to a collection of models which has been trained for per pixel image extension as a Wrapper-Model (WM). Importantly, WM's do not take into account any relationships (positional or otherwise) of the pixel data; each pixel model operates completely independently of all others within a WM. We isolated training and testing image data from each other, and implemented models using scikit Learn without hyper-parameter optimization~\cite{scikit-learn}.

In general, our WM implementation allows for the use of any classification/regression model, though computational feasibility is tightly interwoven with the computational complexity of any underlying model selected. We focused on easily interpretable models. Black and white image data sets were used to train four different types of WM's: decision tree, random forest, perceptron, and linear SVM. Decision tree and random forest are both tree-based algorithms capable of \textit{n}-dimensional decision boundaries, whereas perceptron and linear SVM learn a linear seperating hyperplane with a singular linear decision boundary. Only decision tree WM results are presented for colour images due to model training time constraints.

\begin{figure}[!h]%
\centering
\subfigure[][Dec. Tree]{%
\label{fig:horizontal_dec}%
\fbox{\includegraphics[width=0.23\linewidth]{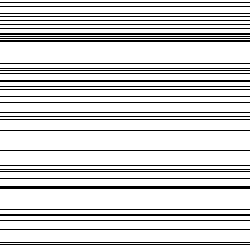}}}%
\hspace{4pt}%
\subfigure[][Ran. Forest]{%
\label{fig:horizontal_ran}%
\fbox{\includegraphics[width=0.23\linewidth]{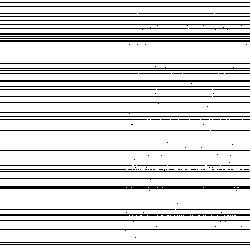}}}%
\hspace{4pt}%
\subfigure[][Perceptron]{%
\label{fig:horizontal_per}%
\fbox{\includegraphics[width=0.23\linewidth]{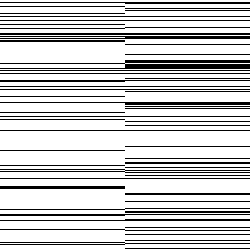}}}%
\hspace{4pt}%
\subfigure[][Lin. SVM]{%
\label{fig:horizontal_lin}%
\fbox{\includegraphics[width=0.23\linewidth]{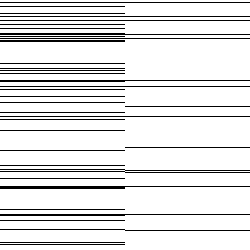}}}%
\vspace{-1em}
\caption{\textbf{Horizontal Lines.} Images created by classification Wrapper-Models trained on 50 randomly generated training images as described in~\Cref{fig:horizontal_input}.}%
\label{fig:horizontal_results}%
\end{figure}

\begin{figure}[!h]%
\centering
\subfigure[][Dec. Tree]{%
\label{fig:vertical_dec}%
\fbox{\includegraphics[width=0.23\linewidth]{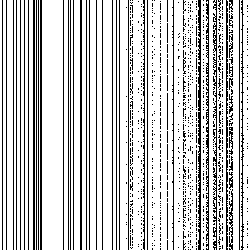}}}%
\hspace{4pt}%
\subfigure[][Ran. Forest]{%
\label{fig:vertical_ran}%
\fbox{\includegraphics[width=0.23\linewidth]{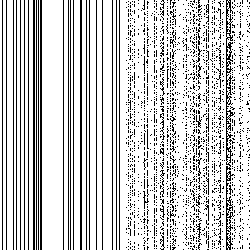}}}%
\hspace{4pt}%
\subfigure[][Perceptron]{%
\label{fig:vertical_per}%
\fbox{\includegraphics[width=0.23\linewidth]{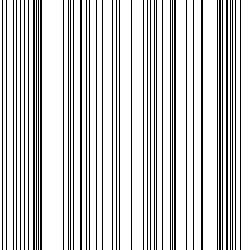}}}%
\hspace{4pt}%
\subfigure[][Lin. SVM]{%
\label{fig:vertical_lin}%
\fbox{\includegraphics[width=0.23\linewidth]{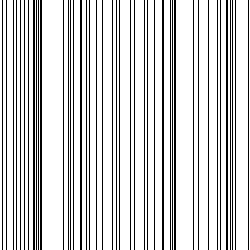}}}%
\vspace{-1em}
\caption{\textbf{Vertical Lines.} Images created by classification Wrapper-Models trained on 50 randomly generated training images as described in~\Cref{fig:vertical_input}.}%
\label{fig:vertical_results}%
\end{figure}

\begin{figure}[!h]%
\centering
\subfigure[][Dec. Tree]{%
\label{fig:circle_dec}%
\fbox{\includegraphics[width=0.23\linewidth]{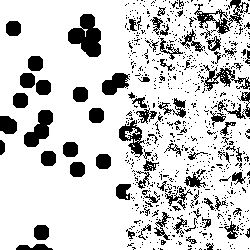}}}%
\hspace{4pt}%
\subfigure[][Ran. Forest]{%
\label{fig:circle_ran}%
\fbox{\includegraphics[width=0.23\linewidth]{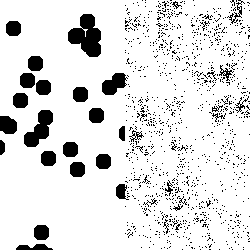}}}%
\hspace{4pt}%
\subfigure[][Perceptron]{%
\label{fig:circle_per}%
\fbox{\includegraphics[width=0.23\linewidth]{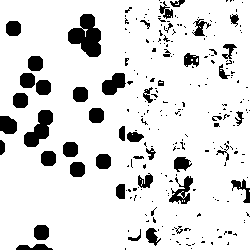}}}%
\hspace{4pt}%
\subfigure[][Lin. SVM]{%
\label{fig:circle_lin}%
\fbox{\includegraphics[width=0.23\linewidth]{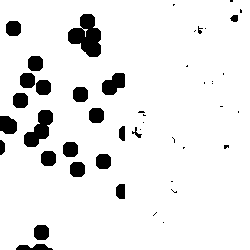}}}%
\vspace{-1em}
\caption{\textbf{Circles.} Images created by classification Wrapper-Models trained on 50 randomly generated training images as described in~\Cref{fig:circle_input}.}%
\label{fig:circle_results}%
\end{figure}

\begin{figure}[!h]%
\centering
\subfigure[][Dec. Tree]{%
\label{fig:ex3-a}%
\fbox{\includegraphics[width=0.23\linewidth]{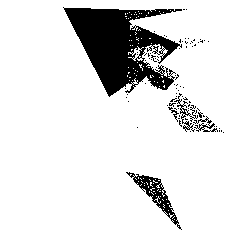}}}%
\hspace{4pt}%
\subfigure[][Ran. Forest]{%
\label{fig:ex3-b}%
\fbox{\includegraphics[width=0.23\linewidth]{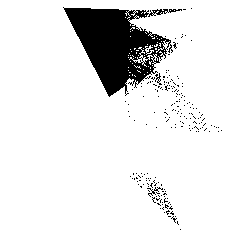}}}%
\hspace{4pt}%
\subfigure[][Perceptron]{%
\label{fig:ex3-c}%
\fbox{\includegraphics[width=0.23\linewidth]{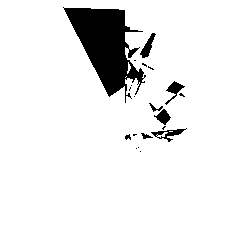}}}%
\hspace{4pt}%
\subfigure[][Lin. SVM]{%
\label{fig:ex3-d}%
\fbox{\includegraphics[width=0.23\linewidth]{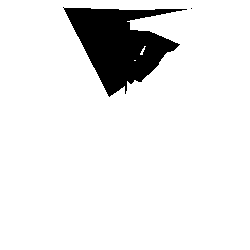}}}%
\vspace{-1em}
\caption{\textbf{Triangle.} Images created by classification Wrapper-Models trained on 50 randomly generated training images as described in~\Cref{fig:triangle_input}.}%
\label{fig:triangle_results}%
\end{figure}

\begin{figure*}[!h]%
\centering
\subfigure[][]{%
\label{final_triangle_a}
\fbox{\includegraphics[width=0.149\linewidth]{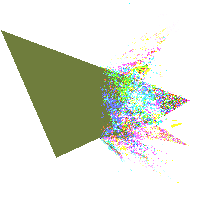}}}%
\subfigure[]{
\label{final_triangle_b}
\fbox{\includegraphics[width=0.149\linewidth]{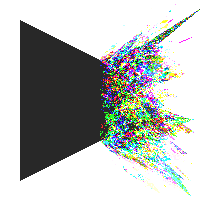}}}%
\subfigure[]{
\label{final_triangle_c}
\fbox{\includegraphics[width=0.149\linewidth]{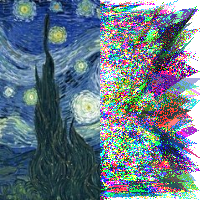}}}%
\subfigure[]{
\label{final_triangle_d}
\fbox{\includegraphics[width=0.149\linewidth]{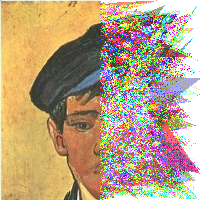}}}%
\subfigure[]{
\label{final_triangle_e}
\fbox{\includegraphics[width=0.149\linewidth]{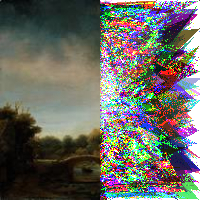}}}%
\subfigure[]{%
\label{final_triangle_f}
\fbox{\includegraphics[width=0.149\linewidth]{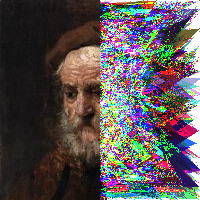}}}%
\vspace{-1.3em}
\caption{\textbf{Triangle WM.} Outputs from a decision tree WM trained on 1900 images of triangles as described in~\Cref{fig:triangle_color_input}}%
\label{fig:final_results_triangle}
\end{figure*}

\begin{figure*}[!h]%
\vspace{-1em}
\centering
\subfigure[][]{%
\label{final_gogh_a}
\fbox{\includegraphics[width=0.149\linewidth]{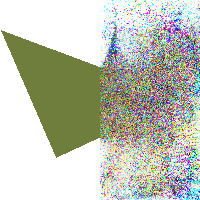}}}%
\subfigure[]{
\label{final_gogh_b}
\fbox{\includegraphics[width=0.149\linewidth]{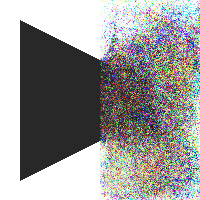}}}%
\subfigure[]{
\label{final_gogh_c}
\fbox{\includegraphics[width=0.149\linewidth]{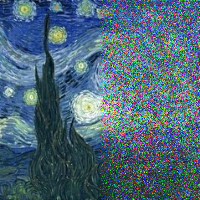}}}%
\subfigure[]{
\label{final_gogh_d}
\fbox{\includegraphics[width=0.149\linewidth]{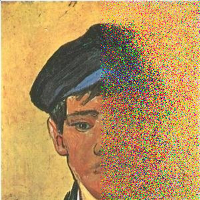}}}%
\subfigure[]{
\label{final_gogh_e}
\fbox{\includegraphics[width=0.149\linewidth]{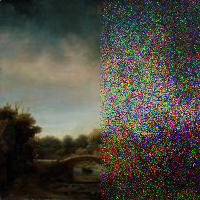}}}%
\subfigure[]{%
\label{final_gogh_f}
\fbox{\includegraphics[width=0.149\linewidth]{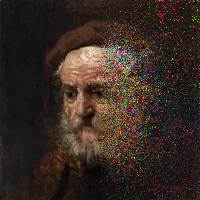}}}%
\vspace{-1.3em}
\caption{\textbf{Vincent van Gogh WM.} Outputs from a decision tree WM trained on 1900 van Gogh works as described in~\Cref{fig:gogh_input}}%
\label{fig:final_results_gogh}
\end{figure*}

\begin{figure*}[!h]%
\vspace{-1em}
\centering
\subfigure[][]{%
\label{final_rem_a}
\fbox{\includegraphics[width=0.149\linewidth]{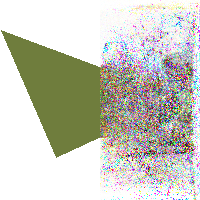}}}%
\subfigure[]{
\label{final_rem_b}
\fbox{\includegraphics[width=0.149\linewidth]{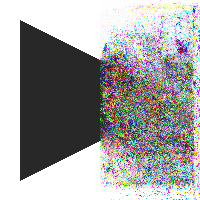}}}%
\subfigure[]{
\label{final_rem_c}
\fbox{\includegraphics[width=0.149\linewidth]{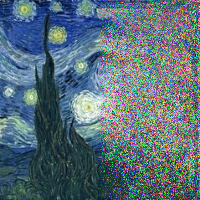}}}%
\subfigure[]{
\label{final_rem_d}
\fbox{\includegraphics[width=0.149\linewidth]{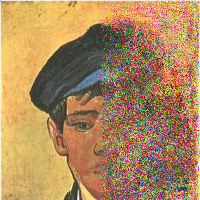}}}%
\subfigure[]{
\label{final_rem_e}
\fbox{\includegraphics[width=0.149\linewidth]{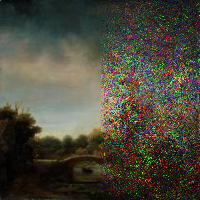}}}%
\subfigure[]{%
\label{final_rem_f}
\fbox{\includegraphics[width=0.149\linewidth]{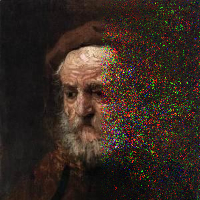}}}%
\vspace{-1.3em}
\caption{\textbf{Rembrandt van Rijn WM.} Outputs from a decision tree WM trained on 900 Rembrandt works as described in~\Cref{fig:rem_input}}%
\label{fig:final_results_rem}
\end{figure*}
\section{Results \& Discussion}

\subsection{Black and White}
Each black and white image type was used to train four WM's, one for each learning approach. Each WM was trained on 50 images\footnote{Black and white images were used as a proof of concept and were therefore trained on small data sets.} and then provided a single left side of the same image type and asked to predict the missing right side. \Cref{fig:horizontal_results,fig:vertical_results,fig:circle_results,fig:triangle_results} present the outputs from each black and white WM for horizontal lines, vertical lines, circles, and triangles respectfully. The left half of each image is the raw input image and the right half is the output prediction from the WM.

Both tree-based WM's were able to correctly predict the horizontal line images (\cref{fig:horizontal_dec,fig:horizontal_ran}), whereas the separating hyperplane WM's were not (\cref{fig:horizontal_per,fig:horizontal_lin}). The opposite is true for the vertical line images where perceptron and linear SVM algorithm outputs look much more natural (\cref{fig:vertical_results}). The accuracy on the vertical line images is far less important than the continuity of the lines which are drawn. Outputs from the circle WM show that none of the models was able to learn the general pattern of circles with the limited number of training images, but each algorithms attempt looks different (\cref{fig:circle_results}). The triangle image WM's all show a general inclination towards convergence, though they differ drastically in their \textit{interpretation} of how to complete the image. More testing with larger training data sets is required to fully interpret how and if the models converge.

\subsection{Colour}
Colour image types were each used to train a single decision tree WM and then used to predict two images of every other colour image type. For example, a decision tree WM was trained on van Gogh images and then provided two image halves from the coloured triangle, van Gogh, and Rembrandt data sets each to extend. In this sense each model is trained on one \textit{style} and then asked to complete images that come from other \textit{styles}. The colored triangle and van Gogh models were trained using 1900 images, and the Rembrandt model was trained using 900 images.
\Cref{fig:final_results_triangle,fig:final_results_gogh,fig:final_results_rem} present the outputs from each of the three trained colour decision tree WM's. As with the black and white images, the left half is the raw input image and the right half is the WM's output.

In all cases each model shows that it is able to predict the general colours and shapes of images in the same style despite never having seen them before. The green triangle WM image shows convergence towards a single point, whereas the second from the left image shows a large spectrum of dark colours with no clear convergence (\cref{final_triangle_a,final_triangle_b}). When tasked with extending van Gogh and Rembrandt works the triangle WM outputs are expectedly abstract (\cref{final_triangle_c,final_triangle_d,final_triangle_e,final_triangle_f}). Van Gogh and Rembrandt WM's are both able to complete images in the style of each other while maintaining a sense of colour and shape; this is especially seen in~\Cref{final_gogh_d,final_gogh_f,final_rem_d,final_rem_f}. When posed with the abstract triangle images the van Gogh and Rembrandt models correctly paint the white background on the fringes and in~\Cref{final_gogh_b} even extend the dark colours from the triangle.

\section{Conclusion}
Through the implementation of a wrapper method which can utilize any single-output classification or regression model to learn from and complete images, Shallow Art presents a novel new method of computational art generation which straddles the boundary between computational creativity and creativity support.

Shallow Art differs from previous approaches in computational creativity support and art generation in its focus on interpretability. This interpretability paves the way for a new analytical perspective which neural network based approaches do not provide. In addition, when framed as an agent in a co-creative process Shallow Art presents an interesting new perspective to discussion on co-creativity, and might enable multi-round experimentation between a human and the Shallow Art system.

Using abstract computer-generated visuals in tandem with artworks from famous classical artists as training data for simple machine learning methods has opened the door to future experimentation and analysis on the topic of non-neural network based approaches to computational art creation.






\bibliographystyle{iccc}
\DeclareRobustCommand{\VAN}[3]{#3}
\bibliography{iccc}

\end{document}